%% file: main.tex
\documentclass[10pt,twocolumn,letterpaper]{article}

\usepackage{amsmath}
\usepackage{adjustbox}

\usepackage[pagenumbers]{cvpr} 
\input{preamble}
\definecolor{cvprblue}{rgb}{0.21,0.49,0.74}
\usepackage[pagebackref,breaklinks,colorlinks,allcolors=cvprblue]{hyperref}

\def\paperID{*****} 
\def\confName{MaCVi}
\def\confYear{2026}

\title{A unified Benchmark for Multi-Frame Image Restoration under Severe Refractive Warping}

\author{
Maxim V. Shugaev$^{1}$, Md Reshad Ul Hoque$^{1}$, Bridget Kennedy$^{1}$, Joseph T. Riley$^{1}$, Fiona Hwang$^{1}$,\\
Justin Hagen$^{2,3}$, Harvir Ghuman$^{2}$, Ethan Garcia-O'Donnell$^{2}$,\\
Syed Noor Qadri$^{2}$, Freddie Santiago$^{2}$, Mun Wai Lee$^{1}$\\[0.4em]
$^{1}$AeroVironment, Inc. \qquad
$^{2}$U. S. Naval Research Laboratory\qquad
$^{3}$George Mason University\\
{\tt\small maxim.shugaev@avinc.com}
}

\begin{document}
\maketitle
\input{sec/0_abstract}    
\input{sec/1_intro}

\input{sec/2_Dataset}

\input{sec/3_experiments}

\input{sec/4_Conclusion}
\input{sec/5_acknowledgments}

{
    \small
    \bibliographystyle{unsrt}
    \bibliography{main}
}

\input{sec/S_suppl}

\end{document}

%% file: sec/0_abstract.tex
\begin{abstract}

Video sequence capturing through refractive dynamic media, such as a turbulent air or water surface, often suffer from severe geometric distortions and temporal instability. While recent advances address mild atmospheric turbulence, no existing benchmarks systematically evaluate restoration methods under strong and highly nonuniform refractive conditions. We present a comprehensive benchmark for geometric distortion removal in video, covering a range from turbulence-like mild warping to strong discontinuous refractive deformations. The benchmark includes both laboratory-captured real data and synthetic sequences generated for static scenes via physics-based light refraction modeling across four distortion levels and multiple surface wave types. We evaluate a spectrum of methods from simple baselines and classical registration algorithms to advanced learning-based approaches including DATUM and our proposed diffusion based V-cache for high and extreme distortions regimes. Evaluation uses both pixel-level (PSNR, SSIM), and perceptual (LPIPS, DINO, CLIP) metrics providing the first large scale analysis of geometric distortion removal. Our benchmark establishes a new foundation for developing and evaluating algorithms capable of reconstructing video from highly distorted optical environments. Our code and datasets are available at \url{https://github.com/iafoss/refractive-mfir-benchmark}.


\end{abstract}

%% file: sec/1_intro.tex
\section{Introduction and Related Work}
\label{sec:intro}

Recovering visually coherent video from imagery degraded by extreme geometric distortions remains a fundamental and unsolved problem in computational imaging. Unlike additive noise, blur, or compression artifacts, these distortions arise from nonlinear refraction phenomena, e.g., atmospheric turbulence, dynamic water surface and Snell’s law, or nonuniform thermal gradients in the transparent hot air, which warp the optical waveform before it reaches the sensor. The result is a sequence of frames that are spatially misaligned and temporary inconsistent, with additional blur introduced by the rapid, time-varying refractive flow during each exposure. Despite a progress in recent years in mitigating turbulence or moderate distortions, there is still no general benchmark capturing a full range of severe spatiotemporal chaotic geometric degradations observed in real environment.

Prior work has addressed the specific subdomains of the problem. Atmospheric turbulence mitigation datasets focus on long-range imaging through air, e.g., CLEAR \cite{anantrasirichai2013atmospheric, anantrasirichai2018atmospheric}, DOST \cite{qin2024unsupervised}, and UG2+ \cite{yuan2019ug}, and provide synthetic and real sequences degraded by mild to moderate distortion. These datasets enable evaluation of alignment methods, e.g. lucky imaging \cite{law2006lucky}, DATUM \cite{zhang2024spatio}, PiRN \cite{jaiswal2023physics}, but largely assume small displacements and smooth temporal evolution. At the other side, work on through-water imaging has explored the effects of dynamic water-air interface (WAI), where wave-induced refraction introduces stronger rapidly varying geometric distortion. Unfortunately, publicly available datasets \cite{thapa2020dynamic, james2019restoration, tian2009seeing} are relatively small and focus on the medium level of distortion. The Sea-Undistort \cite{kromer2025sea} dataset primarily considers color attenuation and scattering rather than the deformation field. In addition, while synthetic data provide paired distorted and undistorted imagery for supervised learning and evaluation, there is a lack of real world videos and there remains a domain gap where the synthetic data fails to capture realistic turbulent wave surfaces, lighting variations, and complex scenes. Outside of these domains, video restoration datasets are primarily focused on video deblurring and optical flow estimation. Consequently, current datasets cannot comprehensively measure the model’s ability to handle large, spatially incoherent, and rapid distortions, leaving a major gap between synthetic turbulence evaluation and the challenges observed in real refractive environments.

\par Previous work for mitigating geometric distortions in video may be grouped into three broad categories: physics-based analytical modeling, multi-frame fusion methods, and learning-based approaches. Analytical approaches attempt to explicitly model the WAI geometry and refraction effect. Jian {et al.} \cite{jian2022seeing} used structured-light projection to estimate the instantaneous water surface geometry and applied reverse ray tracing for single image restoration, which was later refined using Helmholtz-Hodge decomposition of the distortion field \cite{jian2023reconstruction}. Qian {et al.} \cite{qian2018simultaneous} used a camera array to jointly perform 3D reconstruction of the water surface and reconstruct the scene. These analytical approaches require complex hardware setup and calibration. 

\par Multi-frame fusion approaches do not require special hardware but instead use multiple frames to estimate a stable image of a static scene. Zhang {et al.} \cite{zhang2019reconstruction} used a patch-based iterative registration approach combined with blind deconvolution algorithm to construct a reference frame. Jian {et al.} \cite{jian2022water} used global optimization JADE algorithm and a local optimization LBFGS algorithm to perform non-rigid image registration efficiently. Li \textit{et al.} \cite{li2021unsupervised} use an approach inspired by NERF on training a model predicting deformation grids and the undistorted image for a given image sequence. These methods struggle with low texture scenes and strong image distortions. 
\par In the third category, learning-based image restoration methods learn to estimate and reverse image distortions using data driven or deep neural network models. DATUM \cite{zhang2024spatio} blends ideas from classical multi-frame methods as frame alignment, lucky patch fusion, blind deconvolution into a recurrent network architecture that performs temporal aggregation. MambaTM \cite{zhang2025learningphasedistortionselective} expands this approach with selective state space model to expand the receptive field. Thapa \textit{et al.} \cite{thapa2021learning} trained a distortion-guided network (DG-NET) to predict the distortion-free underwater image. Li \textit{et al.} \cite{li2024removing} used a generative adversarial network (GAN) with an attention-based encoder-decoder architecture to remove spatially varying refractive distortions. Such learning-based methods require large training data, which are often simulated, and therefore are limited by the domain gap between synthetic and real-world distortions. Furthermore, since existing benchmarks mostly capture mild to medium levels of distortion, these models are not systematically evaluated in extreme regimes.

\par We introduce a comprehensive benchmark for removal of extreme geometric distortion from a sequence of frames, designed to bridge the gap between atmospheric turbulence mitigation and WAI refractive scenarios. The benchmark focuses on light propagating through dynamic water surfaces, modeling surface perturbations that produce a range of image degradations from mild distortions, comparable to turbulence mitigation, to extreme regimes. These regimes are characterized by discontinuities and multiple apparent instances of the same objects caused by strongly nonlinear image mapping due to multi-path refraction from a dynamic surface. In addition, we propose V-cache, a novel multi-frame diffusion method for effective restoring videos degraded by high and extreme levels of geometric distortions. In summary, our contributions are the following:

\begin{enumerate}
    
    \item {A comprehensive benchmark dataset for extreme geometric distortion removal from video, including both laboratory-captured real sequences and synthetic data generated for static scenes under controlled conditions.}
    \item {A systematically synthetic dataset design that varies distortion amplitude and surface wave type (with surface normal derived from physically based water wave simulation), producing a continuum of refractive conditions from turbulence-like mild warping to severe discontinuous deformations.
    }

    \item{Evaluation covering a wide range of approaches from simple baselines (first-frame and mean-frame fusion), through multi-frame registration (e.g. grid deformation \cite{li2021unsupervised}), to modern learning based models such as DATUM \cite{zhang2024spatio} and our proposed V-cache.}

    \item{A unified evaluation protocol combining pixel-based (PSNR, SSIM) and perception (LPIPS, DINO, CLIP) metrics, allowing rigorous and perceptually consistent assessment.}
\end{enumerate}

%% file: sec/2_Dataset.tex
\section{Dataset}
\label{sec:dataset}

\subsection{Synthetic data}

\begin{figure*}[t]
    \centering
    \includegraphics[width=.7\textwidth]{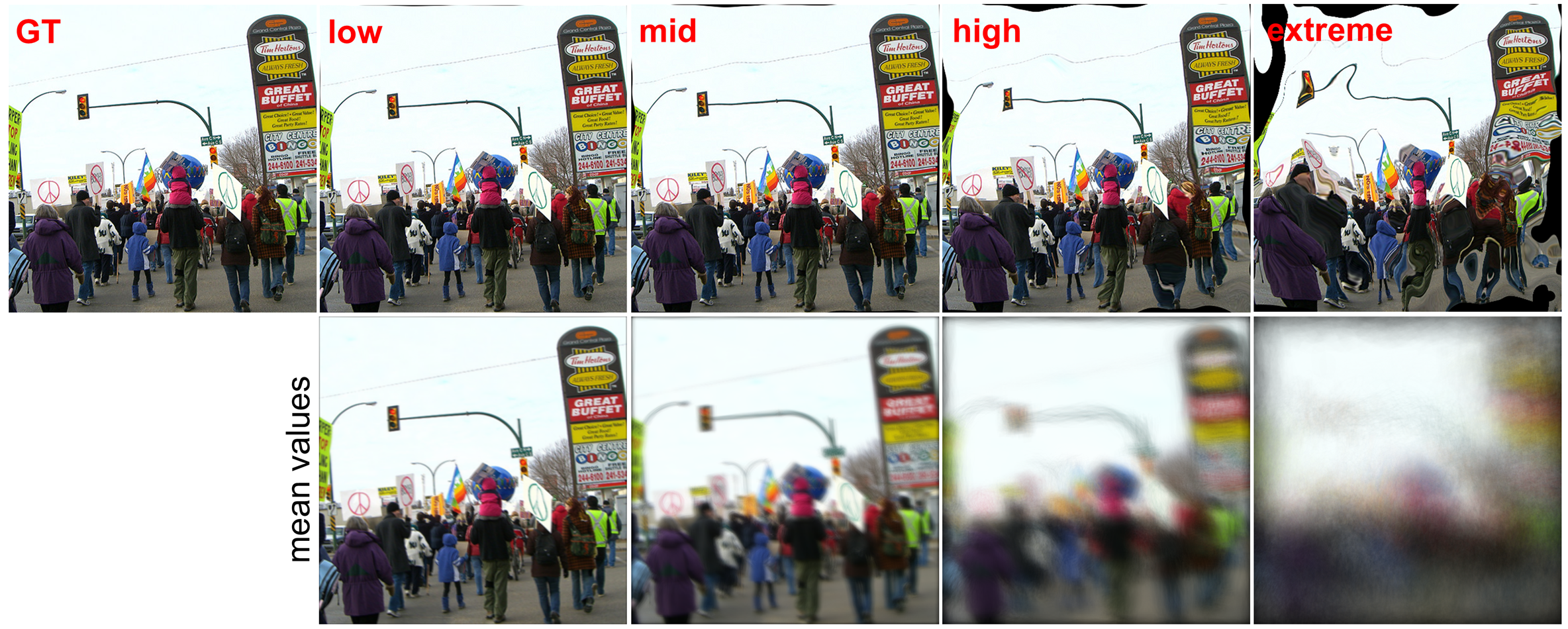}
    \caption{An illustration of the distortion level for different wave amplitudes for ocean waves. Since low and mid amplitudes provide minimal perceptual distortion, we also present the mean pixel values averaged over the entire image sequence.}
    \label{fig:Syn_data}
\end{figure*}

The synthetic data is constructed by combining a particular image background with a wave profile. We consider four distinct types of waves: ocean, shallow water, sine, and ripples with details provided below and in the Supplement. In total, the synthetic evaluation set contains of 30 backgrounds and 10 200-frame-long wave profiles for each wave, and we select a fixed subset of 60 combinations as the final benchmark. We adjust the wave amplitude in the generation process and consider 4 amplitude levels: low, mid, high, and extreme, which correspond to 0.002, 0.006, 0.018, and 0.054 average \textit{std} displacement (relative to the image size) over 10 considered profiles for a given wave type. The speed of waves is selected to maintain a comparable rate of distortion for all wave types. Examples illustrating the corresponding degree of image distortion are shown in {Figure~\ref{fig:Syn_data}. The above combinations produce 16 benchmarks accounting for different wave types and degrease of distortion. The wave profiles, background images, and the code generating the evaluation data are available at ref. \cite{camera_ready_dataset_code, camera_ready_dataset_data}

\subsection{Real data}

The LAB data collection contains a set of undistorted views paired with video recorded in presence of waves. In this set up, a large water tank (20 × 7 × 3 feet) was filled with approximately 19-inch depth of water. A TV monitor was placed above the water to display a set of background images. The camera was set up below the water tank pointing towards the TV. During video recording, a programmable wave generator at one end of the water tank produced disturbance on the water surface with a sine wave profile of a frequency between 1.0 to 2.8 Hz and wave amplitude ranging from 3 to 15 mm peak-to-peak. Besides the sine wave generator, two water pumps were used to generate additional random ripple-like waves. A schematic layout of the data collection hardware is provided in the supplementary material. 

We divide the collected videos into 200 frame clips and resize to 512×512 pixels to have a uniform framework for evaluation. The total number of clips is 216 that combines cases both with low and high levels of distortions.

%% file: sec/3_experiments.tex
\section{Baseline Experiments}

 \begin{figure*}[t]
    \centering
    \includegraphics[width=.65\textwidth]{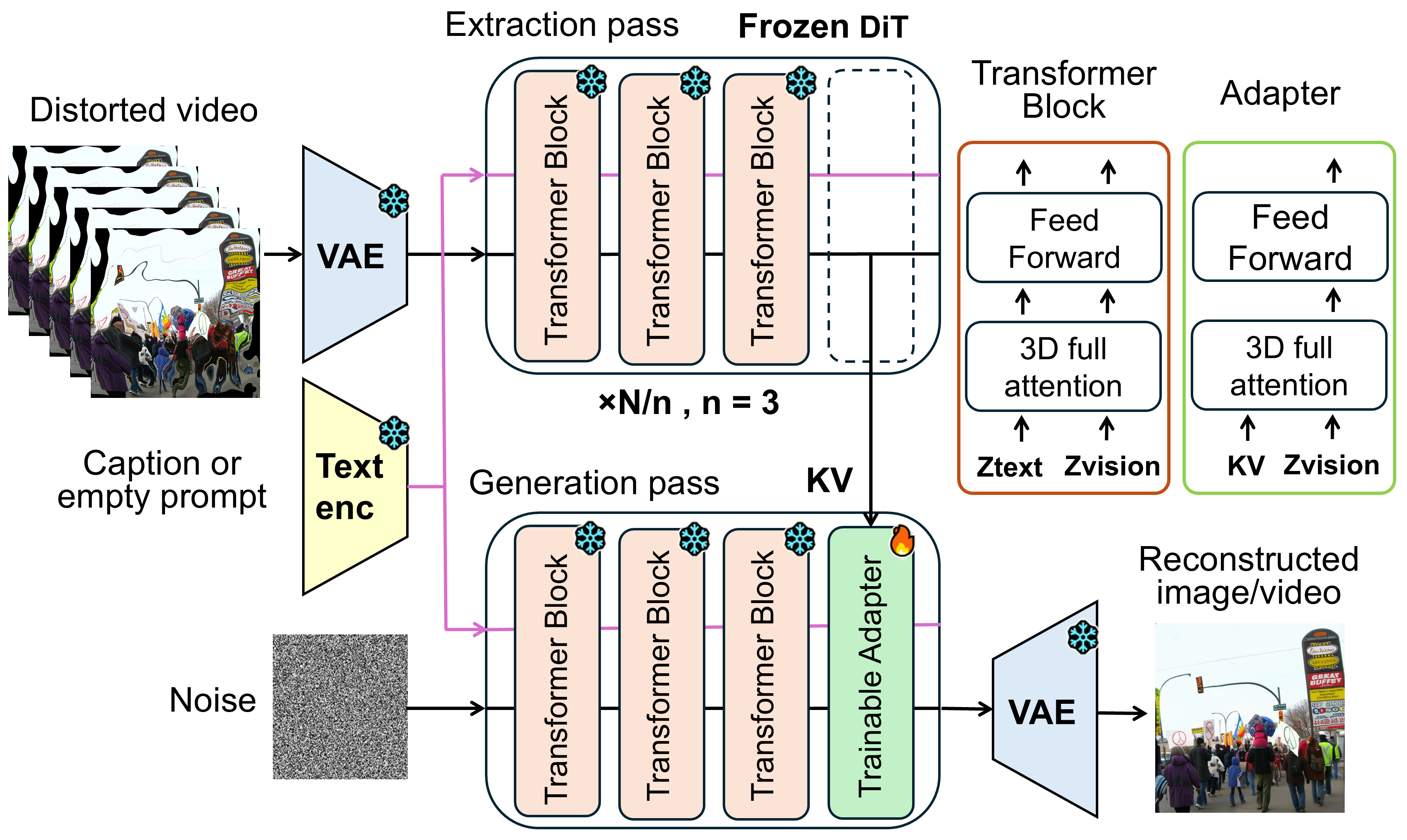}
    \caption{Schematic illustration of V-cache. N refers to the total number of transformer blocks in the model.}
    \label{fig:V-cache}
\end{figure*}

\subsection{Baseline Models}
In our analysis we consider the following approaches. \textit{First frame} – select the first frame of the input video as a prediction, which serves as a reference characterizing the degree of distortion. \textit{Pixel average} – temporally average values of pixels and present it as a prediction.

The next set of models is based on recovery of deformation grids corresponding to the distortion in the input. \textit{Grid deformation} (Non-Rigid Image Distortion Removal) is a self-supervised method described in ref. \cite{li2021unsupervised}. Specifically, given a sequence of distorted frames, the method employs a trainable grid prediction network to estimate the non-rigid deformation field, and a frame reconstruction network that is conditioned on this field to generate the restored image. During training the model learns two complementary mappings: when conditioned on deformed grids, the reconstruction network is supervised to reproduce the corresponding distorted frame, while for a uniform grid, it is expected to produce undistorted (restored) image. The restored output is then warped with predicted deformation field and matched to the input frames. This approach involves training the models for each input sequence and therefore is extremely computational expensive. To keep the inference manageable we downscale input sequence to 256×256 and select only first 11 frames with stride of 2, which reduces the time to approximately 5.2 minutes per sequence. 

Another approach is \textit{Grid registration}, which may be considered as a simplified version of \textit{Grid deformation} where the trainable deformation grids are defined explicitly and applied through warp operation. We assign a coarse learnable grid deformation (1 grid point per 16×16 pixels) to each frame and maximize all vs. all pixel similarity between frames in the video as the optimization objective. Basically this method fits transformations that bring a sequence of distorted frames towards the same appearance, which approximate the static scene. For long sequence, to avoid square increase of the computational complexity, we limit the number of comparisons for each frame by 12 random frames. We apply regularization to prevent large grid deformation or drift. The grid optimization is performed at 256×256 input size, but we warp the original input rather than reduced one with the grid to mitigates blur (which is apparent in \textit{Grid deformation} output) and improve perception metrics. In addition, better computational efficiency enables running this method on the entire 200-frame-long input resulting in the runtime of 39 seconds per sequence.

\par \textit{DATUM} \cite{zhang2024spatio} is a deep-learning based approach designed to remove distortions in a sequence of images caused by atmospheric turbulence. This approach blends ideas from classical multi-frame methods as frame alignment, lucky patch fusion, blind deconvolution into a recurrent network architecture that performs temporal aggregation. Within the network modules like deformable attention alignment and temporal channel attention handle pixel registration and selective fusion, while the twin decoder jointly corrects tilt and blur degradations. In the turbulence problem DATUM archives superior performance and runs several orders of magnitude faster than prior methods taking sub-second time per sequence. Despite this success, the model design limitations, \textit{i.e.} limited receptive field, local feature aggregation, coarse hierarchical flow estimation, and short temporal content aggregation may provide a challenge for modeling rapidly changing strong distortions. Specifically, in our experiments, fine-tuning the model on mid-to-high-amplitude ocean waves caused the appearance of artifacts and degraded performance. Therefore, in our comparison, we use only the original model weights. The inference is performed at 512×512 resolution with 49-frame-long sequences.

\par \textit{V-cache}. Recent advancements of diffusion models, which gradually transform random noise into output through a sequence of denoising steps, have enabled high quality text-to-image and text-to-video generation \cite{ho2020denoising, ramesh2022hierarchical, rombach2022high, blattmann2023stable, liu2024sora}. Public availability of diffusion models \cite{yang2024cogvideox, ramesh2022hierarchical,peebles2023scalable, liu2024sora} has facilitated efforts to enable controllable generation beyond conditioning on text \cite{zhang2023adding,ye2023ip,ruiz2023dreambooth, gal2022image}. The current task of heavy distortion removal may be considered as image or video generation (in case of dynamic scenes) conditioned on another video. Since the output is not pixel aligned with the input, ControlNet-like methods \cite{zhang2023adding} are ineffective. Therefore, as a starting point we considered an encoder-based method: extraction of features from a sequence of frames with DINO \cite{oquab2023dinov2} encoder and finetuning the diffusion model conditioned on these features. Despite the ability of this method to remove the distortion, proper feature alignment is problematic, and reproduction of fine details of the input video, \textit{e.g.} text, is difficult. One of the recent works \cite{aiello2025dreamcache} considers a problem of personalized image generation using a reference-based method conditioning generation directly based on features extracted by the diffusion denoiser. We follow the same logic in the method proposed in our work, {Figure.~\ref{fig:V-cache}. Specifically, first we extract features from the distorted video input at zeroth denoising step and cache them. Then the denoising process is conditioned by these cached features with using adapters inserted into the denoiser ({Figure~\ref{fig:V-cache}}). 

\par As the base model we chose CogVideoX-2B \cite{yang2024cogvideox} because it performs temporal compression in VAE and has full cross-attention between all frames enabling effective input interpretation even under conditions of strong and fast distortions. During training, the base model is frozen, and injection of features is performed with learnable adapters containing cross-attention followed with MLP, as illustrated in {Figure.~\ref{fig:V-cache}. The adapters are inserted after every 5th and 3rd layers, A5 and A3 setups, respectively, which correspond to 180M and 300M trainable parameters in total. 

\par At the generation stage we consider single frame output since the scene is static, but the method is expandable to dynamic scenes with generating a sequence of frames. The total number of processed samples in training is approximately 2.5M, and we dynamically generate samples randomly mixing precomputed wave profiles and backgrounds, as described in the section on synthetic data generation. The model is trained only on ocean wave samples at high wave amplitude range to explore capabilities of this method in removing strong distortions unreachable by registration and DATUM like methods. In contrast to previously published use cases of adapters for reference-based methods \cite{aiello2025dreamcache, hu2024animateanyoneconsistentcontrollable}, where the model extracts an object from one content and places it to another one, in this setup the model is required to reconstruct the undeformed view across multiple frames with high degree of distortion. Therefore, training is relatively long taking 50k steps with 48 batch size. For the 49-frame input setup, the training takes about one month on six A6000 GPUs. We use the native model resolution of 480×720, and at evaluation the input/output is resized from and to 512×512, which may lead to underestimation of the model performance especially in low distortion regime. The inference time is approximately 10 seconds per sequence.

\subsection{Metrics}
We use pixel based and perception metrics to evaluate the quality of image restoration in the considered distortion removal task. PSNR (Peak Signal-to-Noise Ratio) and SSIM (Structural Similarity Index) \cite{wang2004image} are classical reference-based metrics that measure pixel level fidelity. PSNR emphasizes exact intensity matching, rewarding images that resemble the ground truth, while SSIM models track local luminance, contrast, and structural similarity more precisely. These metrics are simple and interpretable but struggle with correct assessment of blurry or slightly misaligned input, failing to reflect human visual quality perception.

\par Perception metrics, e.g. LPIPS (Learned Perceptual Image Patch Similarity) \cite{zhang2018unreasonable}, DINO and CLIP scores \cite{ghildyal2025foundation}, use deep feature embeddings from pretrained networks to assess high level visual similarity. LPIPS measures the differences in intermediate CNN features, which is aligned with human perception of sharpness and texture. DINO and CLIP scores provide one more step forward utilizing high quality foundational models as a feature extractor, providing the best agreement with human perception of quality. Use of transformer-based global representations enables capturing scene level coherence as well. However, these metrics can overlook fine structural errors and may over reward perceptually plausible but geometrically inaccurate outputs. In our work, LPIPS is evaluated using features from the VGG network if not specified explicitly, and we use $L_2$ distance between intermediate features for \textit{dinov1} and \textit{clip\_vitb32} models, which are referred to as DINO and CLIP, respectively.

\subsection{Results and discussion}


\begin{table*}[t]
\centering
\small
\setlength{\tabcolsep}{6pt}
\renewcommand{\arraystretch}{1.12}

\begin{minipage}{\textwidth}
\centering
\captionof{table}{Comparison of PSNR$\uparrow$ for synthetic waves. L, H – low and high wave amplitude. (*) refers to evaluation on multiple output frames and average the metric.}
\label{tab:psnr_compare}
\begin{tabular}{lcccccccc}
\toprule
& \multicolumn{2}{c}{Ocean} & \multicolumn{2}{c}{Shallow water} & \multicolumn{2}{c}{Sine} & \multicolumn{2}{c}{Ripples} \\
\cmidrule(lr){2-3} \cmidrule(lr){4-5} \cmidrule(lr){6-7} \cmidrule(lr){8-9}
Setup & L & H & L & H & L & H & L & H \\
\midrule
First frame        & 23.43 & 14.47 & 21.71 & 13.27 & 21.37 & 13.41 & 20.66 & 13.07 \\
Pixel average      & \textbf{27.69} & 17.40 & \textbf{25.67} & 16.52 & 25.82 & 16.78 & 25.01 & 16.41 \\
Grid deformation   & 25.19 & 15.87 & 20.76 & 14.21 & 20.76 & 14.14 & 21.59 &	14.37 \\
Grid registration* & 25.44 & 15.08 & 24.81 & 13.94 & 25.84 & 14.96 & \textbf{25.93} & 13.12 \\
DATUM*             & 26.19 & 15.60 & 23.43 & 13.66 & \textbf{26.02} & 14.94 & 22.07 & 13.88 \\
V-cache A5       & 23.95 & 21.16 & 23.35 & 17.45 & 22.95 & 20.54 & 23.69 & 17.73 \\
V-cache A3       & 24.86 & \textbf{22.22} & 23.92 & \textbf{17.17} & 23.87 & \textbf{22.07} & 24.02 & \textbf{18.32} \\
\bottomrule
\end{tabular}
\end{minipage}

\vspace{6pt}

\begin{minipage}{\textwidth}
\centering
\captionof{table}{Comparison of LPIPS$\downarrow$ for synthetic waves. L, H – low and high wave amplitude. (*) refers to evaluation on multiple output frames and average the metric.}
\label{tab:lpips_compare}
\begin{tabular}{lcccccccc}
\toprule
& \multicolumn{2}{c}{Ocean} & \multicolumn{2}{c}{Shallow water} & \multicolumn{2}{c}{Sine} & \multicolumn{2}{c}{Ripples} \\
\cmidrule(lr){2-3} \cmidrule(lr){4-5} \cmidrule(lr){6-7} \cmidrule(lr){8-9}
Setup & L & H & L & H & L & H & L & H \\
\midrule
First frame        & 0.075 & 0.356 & 0.099 & 0.503 & 0.099 & 0.380  & 0.129 & 0.521 \\
Pixel average      & 0.172 & 0.608 & 0.204 & 0.612 & 0.196 & 0.521 & 0.214 & 0.556 \\
Grid deformation   & 0.215 & 0.481 & 0.328 & 0.551 & 0.326 & 0.491 & 0.337 &	0.564    \\
Grid registration* & \textbf{0.071} & 0.366 & \textbf{0.076} & 0.464 & \textbf{0.067} & 0.327 & \textbf{0.063} & 0.517 \\
DATUM*             & 0.125 & 0.457 & 0.137 & 0.533 & 0.121 & 0.454 & 0.172 & 0.553 \\
V-cache A5       & 0.134 & 0.186 & 0.135 & 0.256 & 0.152 & 0.190 & 0.142 & 0.323 \\
V-cache A3       & 0.126 & \textbf{0.157} & 0.125 & \textbf{0.247} & 0.153 & \textbf{0.150} & 0.142 & \textbf{0.274} \\
\bottomrule
\end{tabular}
\end{minipage}

\vspace{6pt}

\begin{minipage}{\textwidth}
\centering
\captionof{table}{Comparison of DINO$\downarrow$ for synthetic waves. L, H – low and high wave amplitude. (*) refers to evaluation on multiple output frames and average the metric.}
\label{tab:dino_compare}
\begin{tabular}{lcccccccc}
\toprule
& \multicolumn{2}{c}{Ocean} & \multicolumn{2}{c}{Shallow water} & \multicolumn{2}{c}{Sine} & \multicolumn{2}{c}{Ripples} \\
\cmidrule(lr){2-3} \cmidrule(lr){4-5} \cmidrule(lr){6-7} \cmidrule(lr){8-9}
Setup & L & H & L & H & L & H & L & H \\
\midrule
First frame        & 0.34 & 1.98 & 0.49 & 3.16 & 0.41 & 2.22 & 0.68 & 3.43 \\
Pixel average      & 0.73 & 3.47 & 0.82 & 3.61 & 0.98 & 3.28 & 1.20 & 3.66 \\
Grid deformation   & 0.74 & 2.49 & 1.37 & 3.40 & 1.35 & 2.75 & 1.49 &	3.71   \\
Grid registration* & \textbf{0.30} & 2.07 & \textbf{0.32} & 2.79 & \textbf{0.28} & 1.83 & \textbf{0.26} & 3.34 \\
DATUM*             & 0.40 & 2.53 & 0.42 & 3.31 & 0.38 & 2.52 & 0.59 & 3.59 \\
V-cache A5       & 0.41 & 0.64 & 0.39 & 1.03 & 0.50 & 0.68 & 0.46 & 1.62 \\
V-cache A3       & 0.37 & \textbf{0.45} & 0.36 & \textbf{0.97} & 0.55 & \textbf{0.45} & 0.49 & \textbf{1.23} \\
\bottomrule
\end{tabular}
\end{minipage}

\end{table*}

\begin{figure}
    \centering
    \includegraphics[width=1\linewidth]{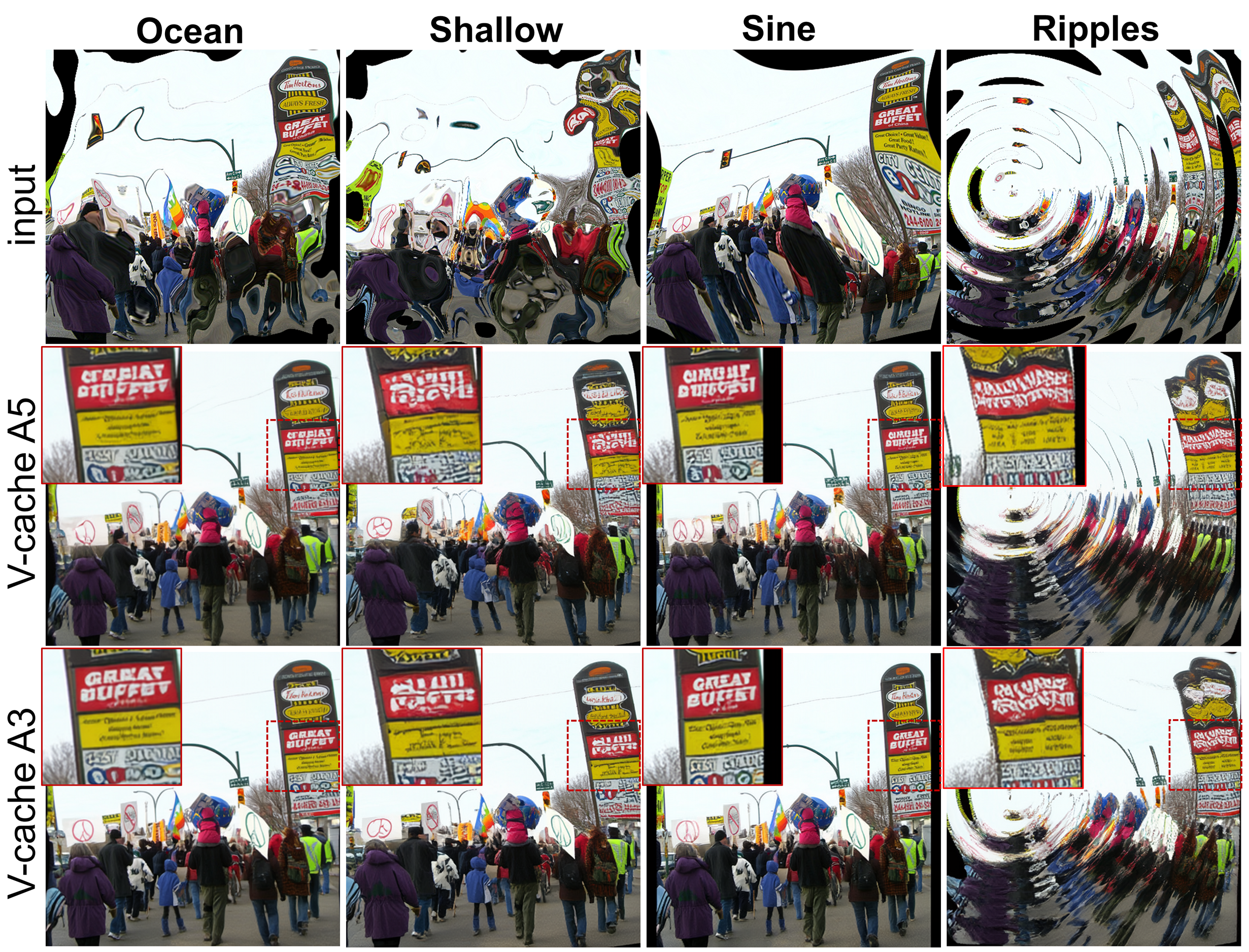}
    \caption{Example of the output produced by V-cache A5 and A3 configurations on the input with extreme distortion level. }
    \label{fig:figure_3}
\end{figure}

\begin{figure}
    \centering
    \includegraphics[width=1\linewidth]{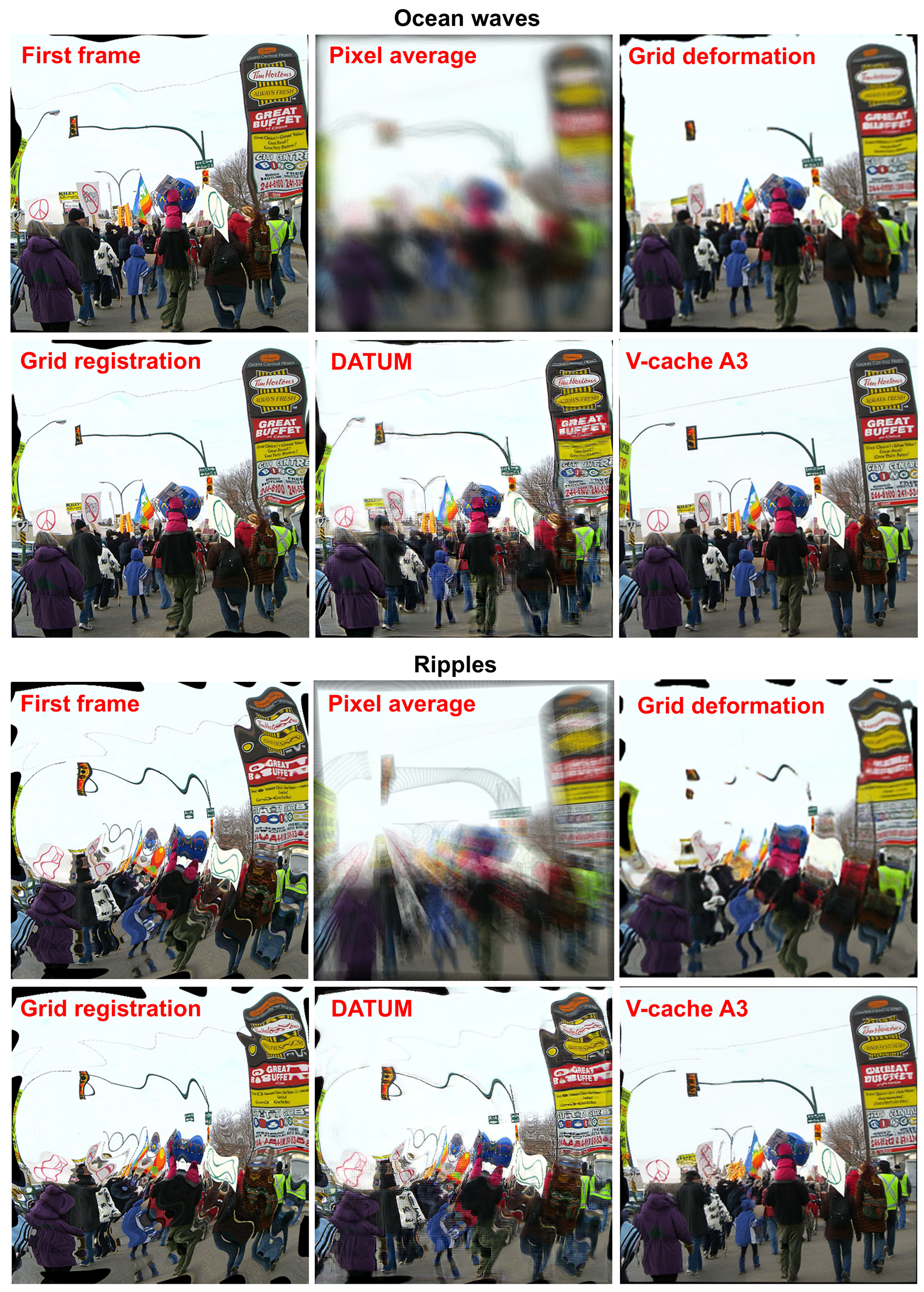}
    \caption{ Example of the output for ocean waves and ripples at high wave amplitude. }
    \label{fig:figure_4}
\end{figure}

\begin{figure}
    \centering
    \includegraphics[width=1\linewidth]{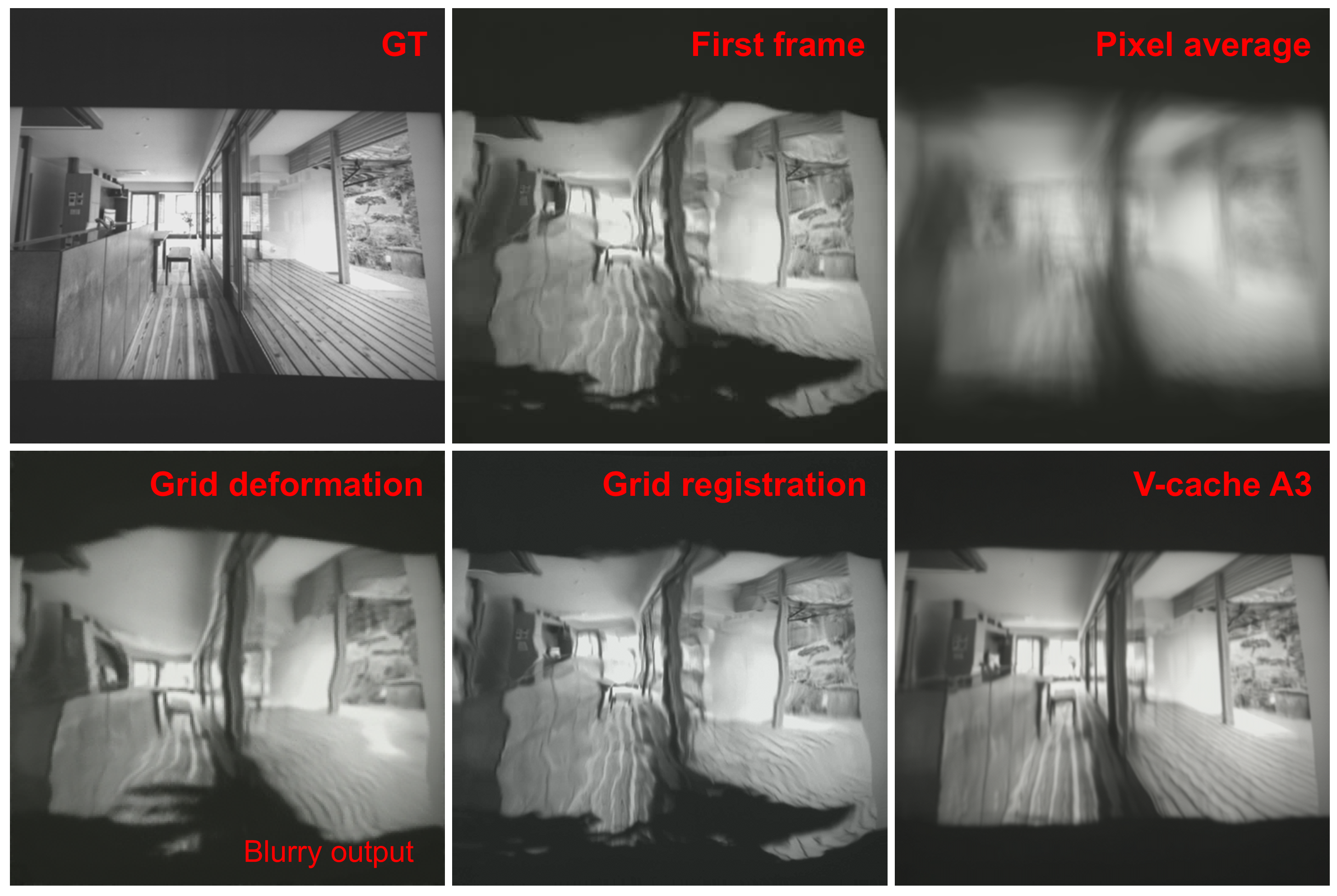}
    \caption{ Example of the output for LAB data with high distortion. }
    \label{fig:figure_5}
\end{figure}

Tables 1-3 summarize the performance of the considered methods on the benchmarks we established. We report results only on low and high wave amplitudes and PSNR, LPIPS, and DINO metrics due to the size limitation, while the full set of results is provided in the Supplement. At the low wave amplitude, comparable to the distortion level introduced by air turbulence, the best results are demonstrated by DATUM and grid registration. Despite DATUM is trained on atmospheric turbulence, it exhibits competitive performance on our benchmark in low distortion case, and further finetuning of the model could boost the performance in this regime. Lower relative performance of V-cache in low distortion regime may be explained by the resizing of the input video and the output image and the training on high wave amplitude only. In addition, exact reproduction of finite image features by this method still may be challenging, which is apparent in low distortion regime. Regarding PSNR, the pixel average benchmark has quite high value, sometimes even outperforming other methods in this regime, while the resulting images are blurry and often visually look worse than the first frame benchmark Figure.~\ref{fig:Syn_data}, which illustrates poor alignment of this metric with human perception.

\par With increase of the wave amplitude the performance of V-cache decreases only moderately compared to other methods, and even in the case of extreme wave amplitude the model still can accurately reproduce the content of the image except ripples, Figure~\ref{fig:figure_3}. Both perception and pixel metrics are dominated by V-cache for a range of wave amplitudes between mid and extreme. Surprisingly, despite the model was not trained on extreme wave amplitude and on waves other than ocean waves, it shows high performance on these out of domain cases and likely could be improved further by training on the specific conditions, especially the case of ripples that currently exhibit difficulties. Figure.~\ref{fig:figure_4} shows a comparison of the output for ocean waves and ripples at high wave amplitude. At this level of distortion, the first frame benchmark has visible deformation, average pixel output looks blurry, and registration-based methods fail. The plot also showcases grid deformation method having blurry output, which is the reason of low values of perception metrics for this method compared to grid registration, despite using similar registration-based concept. Meanwhile, the output of V-cache remains close to the ground truth, and fine details are preserved.

\par Table 4 and Figure.~\ref{fig:figure_5} report results on the LAB evaluation benchmark, which showcase the performance on real data. Since LAB data consists of a combination of cases with various degree of distortion, the difference between reported methods is rather small with V-cache demonstrating the best performance overall. As shown in the plot, this model effectively handles large distortion on real images providing high quality reconstruction.

\begin{table}[t]
\centering
\caption{Comparison of the performance on the LAB dataset. (*) indicates evaluation on a video and averaging the metrics.}
\label{tab:lab_comp}
\small
\setlength{\tabcolsep}{5pt}
\renewcommand{\arraystretch}{1.15}
\begin{adjustbox}{width=\linewidth,center}
\begin{tabular}{lcccc}
\hline
\textbf{Method} & \textbf{PSNR$\uparrow$} & \textbf{SSIM$\uparrow$} & \textbf{LPIPS$\downarrow$} & \textbf{DINO$\downarrow$} \\
\hline
First frame        & 20.6 & 0.740 & 0.330 & 1.80 \\
Pixel average      & 23.3 & 0.792 & 0.379 & 2.25 \\
Grid deformation   & 21.0 & 0.752 & 0.387 & 1.89 \\
Grid registration* & 22.8 & 0.787 & 0.282 & \textbf{0.81} \\
V-cache A5       & 21.1 & 0.755 & 0.349 & 1.80 \\
V-cache A3       & \textbf{23.4} & \textbf{0.831} & \textbf{0.250} & 1.16 \\
\hline
\end{tabular}
\end{adjustbox}
\end{table}

%% file: sec/4_Conclusion.tex
\section{Conclusion}

We introduce a unified benchmark for evaluating multi-frame image restoration under extreme refractive deformation, bridging a gap between atmospheric turbulence mitigation and water-surface refraction scenarios. Our benchmark provides both real and synthetic data, spanning a wide range of distortion severities and wave dynamics, enabling systematic evaluation across multiple physical regimes. In a series of experiments, we benchmarked registration based multi-frame fusion and learning-based approaches, highlighting the strengths and limitations of current methods when faced with highly nonlinear discontinuous refractive warping. The performance of existing methods degrades rapidly as warping intensifies. In contrast, the proposed V-cache is capable of accurate recovery of the input content even under extreme distortion levels, which highlights the ability of learning-based methods to handle this regime. We release our dataset to establish a foundation for future research on video restoration under extreme geometric deformation \cite{camera_ready_dataset_code, camera_ready_dataset_data}.

%% file: sec/5_acknowledgments.tex
\section{Acknowledgments}

This material is based upon work supported by the Naval Research Laboratory (NRL) under Contract No. N00173-24-C-0001. Any opinions, findings and conclusions or recommendations expressed in this material are those of the authors and do not necessarily reflect the views of the Naval Research Laboratory.

%% file: sec/S_suppl.tex
\clearpage
\maketitlesupplementary

\section{LAB setup}
\label{sec:LAB setup}
{Figure.~\ref{fig:lab_setup}. shows the laboratory data collection set up with a large water tank (20 × 7 × 3 feet) which was filled with approximately 19-inch depth of water. A TV monitor was placed above the water to display a set of background images. The camera was set up below the water tank pointing towards the TV. During video recording, a wave generator at one end of the water tank produced disturbance on the water surface with a sine wave profile of a frequency between 1.0 to 2.8 Hz and wave amplitude ranging from 3 to 15mm peak-to-peak. Besides the sine wave generator, two water pumps were used to generate additional random ripple-like waves. 

\begin{figure}
    \centering
    \includegraphics[width=1\linewidth]{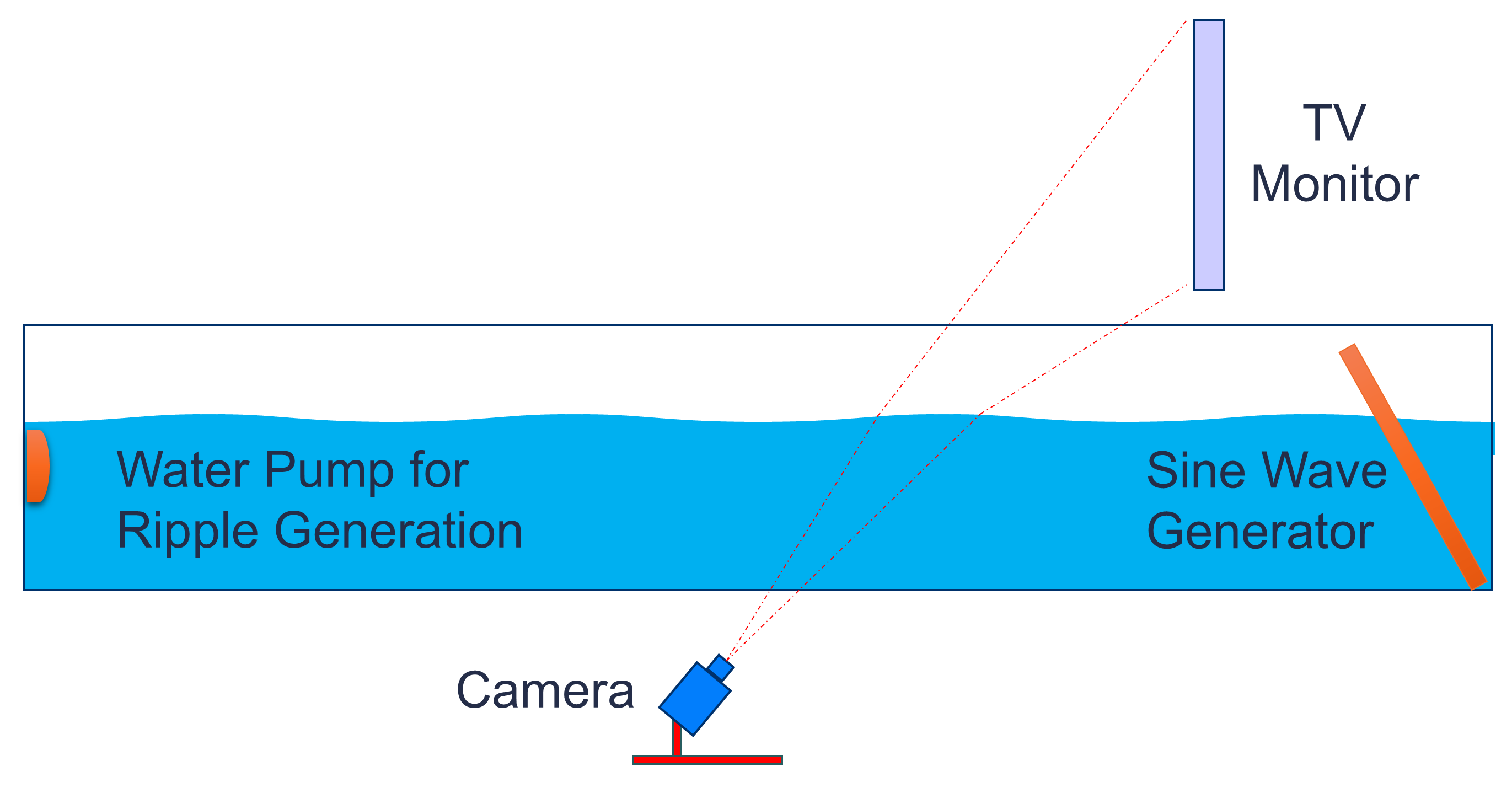}
    \caption{Data collection setup in the lab with water tank and water generators. }
    \label{fig:lab_setup}
\end{figure}

\section{Wave generation}
\label{sec:Wave generation}

Table 5 provides parameters used to generate wave profiles. Details of specific wave types are provided below.

\subsection{Ocean Wave}
For our simulation, we compute the Fast Fourier Transform (FFT) of Gerstner’s equations to represent the wave height as a random field over horizontal position and time. The height $h$($x$,$t$) at the horizontal position  \textbf{x} = ($x$, $z$) can be expressed as

\begin{equation}
  h(\mathbf{x}, t) = \sum_{\mathbf{k}} \tilde{h}(\mathbf{k}, t)\,
  \exp\!\left(i\,\mathbf{k}\cdot\mathbf{x}\right)
\end{equation}

Where $\tilde{h}(\mathbf{k}, t)$
denotes the complex spectral coefficient at wavevector \textbf{k} and time $t$. This spectral form enables efficient analysis and synthesis of the surface, including fine control over amplitudes and phases across the discretized wavenumber domain.  

\par The wave height field is constructed using a spectral model that accounts for wind direction and speed. The Phillips spectrum \cite{tessendorf2001simulating} is used to define the wave amplitude at different wavenumbers. The spectrum is given by:

\begin{equation}
P_{h}(\mathbf{k})
= A \,\frac{\exp\!\left(-\frac{1}{(kL)^{2}}\right)}{k^{4}}
\,\bigl|\hat{\mathbf{k}}\!\cdot\!\hat{\mathbf{w}}\bigr|
\label{eq:phk}
\end{equation}

where $P_h(\mathbf{k})$ is the power spectrum at wave vector $\mathbf{k}$, 
$A$ is an amplitude constant, and for a continuous wind of speed $V$ the
largest attainable scale is $L = V^{2}/g$, where $g$ is gravitational
acceleration and $\hat{\mathbf{w}}$ denotes the unit vector in the wind
direction.

Water-wave height fields can be modeled as Gaussian random fields whose spatial power follows a prescribed spectrum; the most efficient synthesis assigns the corresponding Fourier coefficients and then transforms to physical space.

\begin{equation}
\tilde{h}_0(\mathbf{k}) = \frac{1}{\sqrt{2}}\left(\xi_r + i\,\xi_i\right)\sqrt{P_h(\mathbf{k})}
\end{equation}

The terms $\xi_r$ and $\xi_i$ are two independent random numbers drawn from a standard normal (Gaussian) distribution with mean $0$, variance $1$. After incorporating the Phillips spectrum, conjugate-symmetry, and dispersion, the Fourier amplitudes of the wave field realization ~\cite{tessendorf2001simulating} at time at time $t$ in the frequency domain is:

\begin{equation}
\tilde{h}(\mathbf{k}, t) = \tilde{h}_0(\mathbf{k})\, e^{i\,\omega(\mathbf{k}) t} + \tilde{h}_0^{*}(-\mathbf{k})\, e^{-i\,\omega(\mathbf{k}) t}
\end{equation}

Here, $\tilde{h}_0$ is initial Fourier amplitude (at $t=0$). The height field in Eq.~4 preserves the complex conjugation property by propagating waves ``to the left'' and ``to the right''. To obtain the spatial field $h(\mathbf{x}, t)$, we apply the inverse FFT to Eq.~4 and then compute the spatial gradients of the resulting height map to estimate surface normals and warping displacements. These normals and displacements are subsequently used to render wave-motion effects on the image.

\subsection{Sine Wave}

The sine wave generation model simulates wave surfaces by representing the wave height field as a deterministic sinusoidal function propagating over a 2D spatial domain. This approach simplifies the complex, stochastic nature of waves into a single-frequency wave, suitable for basic visualization or rendering applications. The wave can propagate horizontally with its motion animated over time to mimic wave dynamics. The wave height h, at position x, z and time t (represented by discrete frames) is modeled as a plane wave using a sine function [18]. For image horizontal propagation, the wave height is:              

\begin{equation}
h(x, t) = A \sin\!\left(k_x x - \omega t + \phi\right)
\tag{5}
\end{equation}

where $A$ is the amplitude, $\omega$ is the angular frequency, and $x$ and $y$ are spatial coordinates in a 2D domain, the $\phi$ is the phase offset.

Sine wave is the extended version of the sine wave~\cite{thapa2021learning} to generalize to waves propagating in arbitrary directions by rotating $h(x, t)$ with a random angle.

\subsection{Shallow Water Wave}
The shallow-water equations—derived as a depth-averaged form of the incompressible Navier–Stokes equations \cite{constantin_escher_1998_wavebreaking}—govern conservation of mass and horizontal momentum for a free-surface fluid \cite{thapa2021learning}. Let $h_{sh}$ be the surface height on a Eulerian mesh grid and $(u,v)$ is the 2D velocity, $\rho$ is the fluid density and $g$ is the gravitational acceleration, then the differential equations can be written as

\begin{equation}
\frac{\partial (\rho h_{sh})}{\partial t}
+ \frac{\partial (\rho h_{sh} u)}{\partial x}
+ \frac{\partial (\rho h_{sh} v)}{\partial y}
= 0
\tag{6}
\end{equation}

\begin{equation}
\frac{\partial (\rho h_{sh} u)}{\partial t}
+ \frac{\partial \!\left(\rho h_{sh} u^{2} + \tfrac{1}{2}\rho g h_{sh}^{2}\right)}{\partial x}
+ \frac{\partial (\rho h_{sh} u v)}{\partial y}
= 0
\tag{7}
\end{equation}

\begin{equation}
\frac{\partial (\rho h_{sh} v)}{\partial t}
+ \frac{\partial (\rho h_{sh} u v)}{\partial x}
+ \frac{\partial \!\left(\rho h_{sh} v^{2} + \tfrac{1}{2}\rho g h_{sh}^{2}\right)}{\partial y}
= 0
\tag{8}
\end{equation}

\section{Video generation}
The distorted videos are generated by applying 200-frame-long series of precomputed wave normals to a selected background resized to \(512\times512\). We mimic the LAB setup with camera located underwater and assume low field of view (parallel rays coming from the camera). The vector form of Snell’s law (Eq. 9) is applied to these rays, \(\vec{v}_1\), at the water surface with \(\vec{N}\) surface normals to produce refracted rays \(\vec{v}_2\). \(n1\) and \(n2\) are refractive indexes of water and air, respectively.

\begin{equation*}
\vec{v}_{2}
= \frac{n1}{n2}\, \vec{N} \times \bigl( -\vec{N} \times \vec{v}_{1} \bigr)
  - \vec{N}\, \sqrt{\,1 - \left(\frac{n1}{n2}\right)^{2} \left\lVert \vec{N} \times \vec{v}_{1} \right\rVert_{2}^{2}}
\tag{9}
\end{equation*}

Finally, we compute the lateral displacement of the ray at a given distance to the background from the water surface and generate 2D grid, which then is applied to the background producing the distorted frame. In our experiments we consider four levels of distortion to benchmark model responses in different regimes. We reuse the same wave profiles but scale the degree of deformation both by scaling the distance to the background and surface normals as \(\vec{N}'=(1-\alpha)\cdot\vec{N}_0+\alpha\vec{N}\), where \(\vec{N}_0\) is the vertical normal representing a flat surface. This heuristic enables control of the degree of distortion, and we set the coefficients to ensure average std displacement of \(0.002\), \(0.006\), \(0.018\), and \(0.054\) relative to the image size for low, mid, high, and extreme wave amplitude in all sets regardless the wave type. Since evaluation of refracted rays for each pixel becomes computationally expensive at large image size and long sequence length, we implemented the above procedure on GPU using Pytorch library enabling fast sample generation both in training and inference.

\section{Evaluation on the synthetic data}
Table 5-Table 8 provide a full summary of the evaluation on ocean, shallow water, sine, and ripple waves at low, mid, high, and extreme levels of distortion. Pixel (PSNR and SSIM) and perception metrics (LPIPS, DINO, CLIP) are used. Entire video setup refers to evaluation of the metric for each frame in the video and then averaging. Comparison of this benchmark to the first frame setup may give a clue about variability over the video.


\begin{table*}[t]
\centering
\caption{Evaluation for ocean waves. L, M, H, E — low, medium, high, extreme wave amplitude. (*) refers to evaluation on multiple output frames and average the metric.}
\label{tab:ocean_full}
\small
\setlength{\tabcolsep}{6pt}
\renewcommand{\arraystretch}{1.12}
\begin{tabular}{lcccccc}
\toprule
\textbf{Setup (ocean waves)} & \textbf{PSNR$\uparrow$} & \textbf{SSIM$\uparrow$} & \textbf{LPIPS$_{\text{VGG}}$$\downarrow$} & \textbf{LPIPS$_{\text{Alex}}$$\downarrow$} & \textbf{DINO$\downarrow$} & \textbf{CLIP$\downarrow$} \\
\midrule
\multicolumn{7}{l}{\textit{First frame}} \\
\quad L & 23.43 & 0.813 & 0.075 & 0.016 & 0.339 & 0.066 \\
\quad M & 17.92 & 0.585 & 0.186 & 0.097 & 0.827 & 0.177 \\
\quad H & 14.47 & 0.437 & 0.356 & 0.237 & 1.981 & 0.416 \\
\quad E & 11.76 & 0.343 & 0.527 & 0.457 & 3.462 & 0.749 \\
\midrule
\multicolumn{7}{l}{\textit{Entire video*}} \\
\quad L & 21.78 & 0.754 & 0.096 & 0.055 & 0.432 & 0.086 \\
\quad M & 16.71 & 0.524 & 0.227 & 0.125 & 1.077 & 0.227 \\
\quad H & 13.39 & 0.389 & 0.425 & 0.310 & 2.515 & 0.530 \\
\quad E & 10.88 & 0.305 & 0.586 & 0.544 & 3.980 & 0.888 \\
\midrule
\multicolumn{7}{l}{\textit{Pixel average}} \\
\quad L & 27.69 & 0.866 & 0.172 & 0.199 & 0.729 & 0.160 \\
\quad M & 21.26 & 0.628 & 0.417 & 0.442 & 2.251 & 0.490 \\
\quad H & 17.40 & 0.469 & 0.608 & 0.628 & 3.606 & 0.793 \\
\quad E & 14.71 & 0.417 & 0.660 & 0.735 & 3.824 & 0.862 \\
\midrule
\multicolumn{7}{l}{\textit{Grid deformation}} \\
\quad L & 25.19 & 0.813 & 0.215 & 0.182 & 0.736 & 0.205 \\
\quad M & 19.51 & 0.608 & 0.367 & 0.360 & 1.664 & 0.405 \\
\quad H & 15.87 & 0.481 & 0.481 & 0.231 & 2.493 & 0.560 \\
\quad E & 12.70 & 0.387 & 0.590 & 0.615 & 3.778 & 0.861 \\
\midrule
\multicolumn{7}{l}{\textit{Grid registration*}} \\
\quad L & 25.44 & 0.874 & 0.071 & 0.042 & 0.301 & 0.060 \\
\quad M & 20.65 & 0.721 & 0.144 & 0.079 & 0.655 & 0.137 \\
\quad H & 15.08 & 0.466 & 0.366 & 0.251 & 2.067 & 0.454 \\
\quad E & 11.39 & 0.324 & 0.572 & 0.520 & 3.842 & 0.870 \\
\midrule
\multicolumn{7}{l}{\textit{DATUM*}} \\
\quad L & 26.19 & 0.863 & 0.125 & 0.078 & 0.396 & 0.089 \\
\quad M & 21.09 & 0.718 & 0.229 & 0.147 & 0.902 & 0.198 \\
\quad H & 15.60 & 0.468 & 0.457 & 0.353 & 2.533 & 0.533 \\
\quad E & 12.05 & 0.312 & 0.621 & 0.584 & 4.023 & 0.842 \\
\midrule
\multicolumn{7}{l}{\textit{V-cache A5}} \\
\quad L & 23.95 & 0.789 & 0.134 & 0.077 & 0.045 & 0.407 \\
\quad M & 23.46 & 0.772 & 0.138 & 0.077 & 0.397 & 0.120 \\
\quad H & 21.16 & 0.672 & 0.186 & 0.100 & 0.640 & 0.171 \\
\quad E & 16.54 & 0.502 & 0.331 & 0.207 & 1.594 & 0.372 \\
\midrule
\multicolumn{7}{l}{\textit{V-cache A3}} \\
\quad L & 24.86 & 0.813 & 0.126 & 0.075 & 0.373 & 0.114 \\
\quad M & 24.27 & 0.798 & 0.120 & 0.073 & 0.289 & 0.092 \\
\quad H & 22.22 & 0.717 & 0.157 & 0.089 & 0.449 & 0.129 \\
\quad E & 18.11 & 0.559 & 0.157 & 0.268 & 1.102 & 0.263 \\
\bottomrule
\end{tabular}
\end{table*}

\begin{table*}[t]
\centering
\caption{Evaluation for shallow water waves. L, M, H, E — low, medium, high, extreme wave amplitude. (*) refers to evaluation on multiple output frames and average the metric.}
\label{tab:shallow_full}
\small
\setlength{\tabcolsep}{6pt}
\renewcommand{\arraystretch}{1.12}
\begin{tabular}{lcccccc}
\toprule
\textbf{Setup (shallow water waves)} & \textbf{PSNR$\uparrow$} & \textbf{SSIM$\uparrow$} & \textbf{LPIPS$_{\text{VGG}}$$\downarrow$} & \textbf{LPIPS$_{\text{Alex}}$$\downarrow$} & \textbf{DINO$\downarrow$} & \textbf{CLIP$\downarrow$} \\
\midrule
\multicolumn{7}{l}{\textit{First frame}} \\
\quad L & 21.71 & 0.733 & 0.099 & 0.059 & 0.489 & 0.101 \\
\quad M & 16.65 & 0.492 & 0.264 & 0.149 & 1.370 & 0.296 \\
\quad H & 13.27 & 0.374 & 0.503 & 0.399 & 3.160 & 0.692 \\
\quad E & 10.84 & 0.292 & 0.610 & 0.580 & 4.244 & 0.899 \\
\midrule
\multicolumn{7}{l}{\textit{Entire video*}} \\
\quad L & 20.68 & 0.682 & 0.113 & 0.064 & 0.512 & 0.108 \\
\quad M & 16.07 & 0.462 & 0.273 & 0.150 & 1.384 & 0.293 \\
\quad H & 12.92 & 0.362 & 0.501 & 0.392 & 3.171 & 0.682 \\
\quad E & 10.52 & 0.283 & 0.616 & 0.593 & 4.345 & 0.936 \\
\midrule
\multicolumn{7}{l}{\textit{Pixel average}} \\
\quad L & 25.67 & 0.800 & 0.204 & 0.228 & 0.821 & 0.197 \\
\quad M & 20.03 & 0.536 & 0.446 & 0.427 & 2.468 & 0.509 \\
\quad H & 16.52 & 0.432 & 0.612 & 0.606 & 3.606 & 0.793 \\
\quad E & 14.08 & 0.404 & 0.671 & 0.750 & 3.921 & 0.875 \\
\midrule
\multicolumn{7}{l}{\textit{Grid deformation}} \\
\quad L & 20.76 & 0.648 & 0.328 & 0.326 & 1.370 & 0.354 \\
\quad M & 17.29 & 0.496 & 0.413 & 0.386 & 1.991 & 0.466 \\
\quad H & 14.21 & 0.413 & 0.551 & 0.525 & 3.401 & 0.735 \\
\quad E & 11.64 & 0.352 & 0.636 & 0.668 & 4.377 & 0.937 \\
\midrule
\multicolumn{7}{l}{\textit{Grid registration*}} \\
\quad L & 24.81 & 0.846 & 0.076 & 0.045 & 0.318 & 0.068 \\
\quad M & 19.51 & 0.646 & 0.180 & 0.096 & 0.855 & 0.183 \\
\quad H & 13.94 & 0.397 & 0.464 & 0.336 & 2.791 & 0.628 \\
\quad E & 10.88 & 0.297 & 0.607 & 0.575 & 4.246 & 0.925 \\
\midrule
\multicolumn{7}{l}{\textit{DATUM*}} \\
\quad L & 23.43 & 0.806 & 0.137 & 0.026 & 0.420 & 0.098 \\
\quad M & 19.51 & 0.646 & 0.289 & 0.174 & 1.225 & 0.275 \\
\quad H & 13.66 & 0.374 & 0.533 & 0.434 & 3.306 & 0.689 \\
\quad E & 11.24 & 0.295 & 0.640 & 0.634 & 4.435 & 0.920 \\
\midrule
\multicolumn{7}{l}{\textit{V-cache A5}} \\
\quad L & 23.35 & 0.772 & 0.135 & 0.078 & 0.394 & 0.119 \\
\quad M & 20.85 & 0.673 & 0.165 & 0.090 & 0.521 & 0.151 \\
\quad H & 17.45 & 0.514 & 0.256 & 0.137 & 1.025 & 0.258 \\
\quad E & 13.89 & 0.394 & 0.454 & 0.321 & 2.613 & 0.579 \\
\midrule
\multicolumn{7}{l}{\textit{V-cache A3}} \\
\quad L & 23.92 & 0.788 & 0.125 & 0.077 & 0.359 & 0.110 \\
\quad M & 20.83 & 0.674 & 0.151 & 0.090 & 0.428 & 0.127 \\
\quad H & 17.17 & 0.507 & 0.247 & 0.140 & 0.966 & 0.241 \\
\quad E & 14.17 & 0.406 & 0.415 & 0.280 & 2.220 & 0.496 \\
\bottomrule
\end{tabular}
\end{table*}

\begin{table*}[t]
\centering
\caption{Evaluation for sine waves. L, M, H, E — low, medium, high, extreme wave amplitude. (*) refers to evaluation on multiple output frames and average the metric.}
\label{tab:sine_full}
\small
\setlength{\tabcolsep}{6pt}
\renewcommand{\arraystretch}{1.12}
\begin{tabular}{lcccccc}
\toprule
\textbf{Setup (sine waves)} & \textbf{PSNR$\uparrow$} & \textbf{SSIM$\uparrow$} & \textbf{LPIPS$_{\text{VGG}}$$\downarrow$} & \textbf{LPIPS$_{\text{Alex}}$$\downarrow$} & \textbf{DINO$\downarrow$} & \textbf{CLIP$\downarrow$} \\
\midrule
\multicolumn{7}{l}{\textit{First frame}} \\
\quad L & 21.37 & 0.716 & 0.099 & 0.055 & 0.413 & 0.085 \\
\quad M & 16.61 & 0.507 & 0.213 & 0.115 & 0.956 & 0.209 \\
\quad H & 13.41 & 0.399 & 0.380 & 0.276 & 2.223 & 0.455 \\
\quad E & 10.90 & 0.320 & 0.544 & 0.501 & 3.639 & 0.804 \\
\midrule
\multicolumn{7}{l}{\textit{Entire video*}} \\
\quad L & 21.38 & 0.716 & 0.099 & 0.055 & 0.417 & 0.086 \\
\quad M & 16.64 & 0.506 & 0.214 & 0.312 & 0.962 & 0.209 \\
\quad H & 13.42 & 0.397 & 0.381 & 0.278 & 2.238 & 0.452 \\
\quad E & 10.81 & 0.315 & 0.547 & 0.507 & 3.683 & 0.808 \\
\midrule
\multicolumn{7}{l}{\textit{Pixel average}} \\
\quad L & 25.82 & 0.809 & 0.196 & 0.172 & 0.984 & 0.431 \\
\quad M & 20.20 & 0.568 & 0.383 & 0.065 & 2.217 & 1.777 \\
\quad H & 16.78 & 0.446 & 0.521 & 0.468 & 3.285 & 3.110 \\
\quad E & 14.18 & 0.398 & 0.618 & 0.636 & 3.946 & 4.039 \\
\midrule
\multicolumn{7}{l}{\textit{Grid deformation}} \\
\quad L & 20.76 & 0.649 & 0.326 & 0.323 & 1.349 & 0.351 \\
\quad M & 17.24 & 0.505 & 0.393 & 0.366 & 1.769 & 0.423 \\
\quad H & 14.14 & 0.417 & 0.491 & 0.458 & 2.746 & 0.584 \\
\quad E & 11.64 & 0.353 & 0.595 & 0.609 & 3.942 & 0.842 \\
\midrule
\multicolumn{7}{l}{\textit{Grid registration*}} \\
\quad L & 25.84 & 0.872 & 0.067 & 0.040 & 0.287 & 0.054 \\
\quad M & 21.41 & 0.739 & 0.113 & 0.149 & 0.478 & 0.103 \\
\quad H & 14.96 & 0.457 & 0.327 & 0.217 & 1.831 & 0.383 \\
\quad E & 11.20 & 0.324 & 0.542 & 0.490 & 3.629 & 0.811 \\
\midrule
\multicolumn{7}{l}{\textit{DATUM*}} \\
\quad L & 26.02 & 0.855 & 0.121 & 0.077 & 0.380 & 0.084 \\
\quad M & 19.68 & 0.644 & 0.245 & 0.087 & 0.937 & 0.209 \\
\quad H & 14.94 & 0.441 & 0.454 & 0.338 & 2.515 & 0.521 \\
\quad E & 11.65 & 0.319 & 0.613 & 0.575 & 3.994 & 0.840 \\
\midrule
\multicolumn{7}{l}{\textit{V-cache A5}} \\
\quad L & 22.95 & 0.747 & 0.152 & 0.091 & 0.501 & 0.190 \\
\quad M & 22.68 & 0.734 & 0.150 & 0.312 & 0.467 & 0.137 \\
\quad H & 20.54 & 0.640 & 0.190 & 0.105 & 0.680 & 0.176 \\
\quad E & 15.70 & 0.489 & 0.343 & 0.237 & 1.834 & 0.399 \\
\midrule
\multicolumn{7}{l}{\textit{V-cache A3}} \\
\quad L & 23.87 & 0.769 & 0.153 & 0.093 & 0.545 & 0.153 \\
\quad M & 23.12 & 0.756 & 0.141 & 0.086 & 0.430 & 0.126 \\
\quad H & 22.07 & 0.715 & 0.150 & 0.091 & 0.448 & 0.128 \\
\quad E & 16.70 & 0.532 & 0.288 & 0.198 & 1.340 & 0.304 \\
\bottomrule
\end{tabular}
\end{table*}

\begin{table*}[t]
\centering
\caption{Evaluation for ripples waves. L, M, H, E — low, medium, high, extreme wave amplitude. (*) refers to evaluation on multiple output frames and average the metric.}
\label{tab:ripples_full}
\small
\setlength{\tabcolsep}{6pt}
\renewcommand{\arraystretch}{1.12}
\begin{tabular}{lcccccc}
\toprule
\textbf{Setup (ripples)} & \textbf{PSNR$\uparrow$} & \textbf{SSIM$\uparrow$} & \textbf{LPIPS$_{\text{VGG}}$$\downarrow$} & \textbf{LPIPS$_{\text{Alex}}$$\downarrow$} & \textbf{DINO$\downarrow$} & \textbf{CLIP$\downarrow$} \\
\midrule
\multicolumn{7}{l}{\textit{First frame}} \\
\quad L & 20.66 & 0.712 & 0.129 & 0.077 & 0.678 & 0.138 \\
\quad M & 16.12 & 0.502 & 0.327 & 0.218 & 1.865 & 0.418 \\
\quad H & 13.07 & 0.384 & 0.521 & 0.430 & 3.434 & 0.746 \\
\quad E & 10.49 & 0.259 & 0.622 & 0.604 & 4.500 & 0.913 \\
\midrule
\multicolumn{7}{l}{\textit{Entire video*}} \\
\quad L & 20.71 & 0.712 & 0.127 & 0.076 & 0.670 & 0.137 \\
\quad M & 16.15 & 0.502 & 0.326 & 0.217 & 1.688 & 0.387 \\
\quad H & 13.06 & 0.383 & 0.521 & 0.429 & 3.345 & 0.740 \\
\quad E & 10.49 & 0.260 & 0.621 & 0.602 & 4.489 & 0.910 \\
\midrule
\multicolumn{7}{l}{\textit{Pixel average}} \\
\quad L & 25.01 & 0.806 & 0.214 & 0.185 & 1.203 & 0.222 \\
\quad M & 19.69 & 0.564 & 0.400 & 0.326 & 2.459 & 0.481 \\
\quad H & 16.41 & 0.443 & 0.556 & 0.485 & 3.665 & 0.730 \\
\quad E & 13.86 & 0.360 & 0.642 & 0.678 & 4.481 & 0.885 \\
\midrule
\multicolumn{7}{l}{\textit{Grid deformation}} \\
\quad L & 21.59 & 0.682 & 0.337 & 0.339 & 1.485 & 0.376 \\
\quad M & 18.24 & 0.555 & 0.433 & 0.420 & 2.379 & 0.541 \\
\quad H & 14.37 & 0.428 & 0.564 & 0.561 & 3.711 & 0.795 \\
\quad E & 11.36 & 0.321 & 0.641 & 0.683 & 4.678 & 0.936 \\
\midrule
\multicolumn{7}{l}{\textit{Grid registration*}} \\
\quad L & 25.93 & 0.888 & 0.063 & 0.039 & 0.263 & 0.056 \\
\quad M & 16.89 & 0.533 & 0.313 & 0.196 & 1.688 & 0.387 \\
\quad H & 13.12 & 0.380 & 0.517 & 0.413 & 3.345 & 0.740 \\
\quad E & 11.02 & 0.310 & 0.571 & 0.523 & 3.953 & 0.857 \\
\midrule
\multicolumn{7}{l}{\textit{DATUM*}} \\
\quad L & 22.07 & 0.753 & 0.172 & 0.103 & 0.591 & 0.136 \\
\quad M & 17.49 & 0.549 & 0.362 & 0.252 & 1.858 & 0.403 \\
\quad H & 13.88 & 0.389 & 0.553 & 0.471 & 3.593 & 0.752 \\
\quad E & 11.35 & 0.258 & 0.661 & 0.681 & 4.629 & 0.923 \\
\midrule
\multicolumn{7}{l}{\textit{V-cache A5}} \\
\quad L & 23.69 & 0.778 & 0.142 & 0.081 & 0.458 & 0.133 \\
\quad M & 21.94 & 0.708 & 0.168 & 0.091 & 0.567 & 0.160 \\
\quad H & 17.73 & 0.529 & 0.323 & 0.178 & 1.616 & 0.380 \\
\quad E & 12.40 & 0.332 & 0.576 & 0.527 & 4.127 & 0.885 \\
\midrule
\multicolumn{7}{l}{\textit{V-cache A3}} \\
\quad L & 24.02 & 0.784 & 0.142 & 0.085 & 0.492 & 0.138 \\
\quad M & 22.79 & 0.741 & 0.148 & 0.083 & 0.460 & 0.132 \\
\quad H & 18.32 & 0.567 & 0.274 & 0.151 & 1.235 & 0.281 \\
\quad E & 12.96 & 0.361 & 0.538 & 0.462 & 3.705 & 0.775 \\
\bottomrule
\end{tabular}
\end{table*}